\documentclass[journal]{IEEEtran}
\usepackage{cite}
\usepackage{graphicx}
\usepackage{comment}
\usepackage{amsmath,amssymb} 
\usepackage{color}

\ifCLASSINFOpdf
\else
\fi
\begin{document}
%
\title{Unifying Graph Embedding Features with Graph Convolutional Networks for Skeleton-based Action Recognition}
%
%
%
\author{Dong Yang\IEEEauthorrefmark{1},
        Monica Mengqi Li\IEEEauthorrefmark{1},
        Hong Fu,
        Jicong Fan,
        Zhao Zhang,
        Howard Leung
\thanks{Dong Yang is with Management of Complex Systems Department, University of California, Merced, USA (dongyang3-c@my.cityu.edu.hk).}
\thanks{Monica Mengqi Li is with Department of Computer Science, Bishop's University, Sherbrooke, Canada (mli@cs.ubishops.ca).}
\thanks{Hong Fu (corresponding author) is with Department of Mathematics and
Information Technology, The Education University of Hong Kong, Hong Kong,
China (hfu@eduhk.hk).}
\thanks{Jicong Fan (corresponding author) is with Department of Computer Science, City University of Hong Kong, Shenzhen, China (fanjicong@cuhk.edu.cn).}
\thanks{Zhao Zhang (corresponding author) is with Key Laboratory of Knowledge Engineering with Big Data (Ministry of Education), School of Computer Science and Information Engineering, Hefei University of Technology (HFUT), Hefei 230601, China (cszzhang@gmail.com)}
\thanks{Howard Leung is with Department of Computer Science, City University of Hong Kong, Hong Kong, China (howard@cityu.edu.hk).}
\thanks{\IEEEauthorrefmark{1} These authors contributed equally}

}

%
%

\markboth{Journal of \LaTeX\ Class Files,~Vol.~14, No.~8, August~2021}%
{Shell \MakeLowercase{\textit{et al.}}: Bare Demo of IEEEtran.cls for IEEE Journals}
%



\maketitle

\begin{abstract}
Combining skeleton structure with graph convolutional networks has achieved remarkable performance in human action recognition. Since current
research focuses on designing basic graph for representing skeleton data, these embedding features contain basic topological information, which cannot learn more systematic perspectives from skeleton data. In this paper, we overcome this limitation by proposing a novel framework, which unifies 15 graph embedding features into the graph convolutional network for human action recognition, aiming to best take advantage of graph information to distinguish key joints, bones, and body parts in human action, instead of being exclusive to a single feature or domain. Additionally, we fully investigate how to find the best graph features of skeleton structure for improving human action recognition. Besides, the topological information of the skeleton sequence is explored to further enhance the performance in a multi-stream framework. Moreover, the unified graph features are extracted by the adaptive methods on the training process, which further yields improvements. Our model is validated by three large-scale datasets, namely NTU-RGB+D, Kinetics and SYSU-3D, and outperforms the state-of-the-art methods. Overall, our work unified graph embedding features to promotes systematic research on human action recognition.
\end{abstract}


%
\IEEEpeerreviewmaketitle

\section{Introduction}
%
%
%
%
\IEEEPARstart{H}{uman} action recognition has attracted substantial attention from different research areas in recent years, since the deluge of skeleton data offers unprecedented opportunities to investigate structures, appearance, joints and bones, video sequences of human actions \cite{yan2018spatial,luo2019tangent,song2017end}. The study of human action recognition provides significant insights for action surveillance, pose tracking, medical monitoring and diagnosis, sports analysis, security reports, and human-computer interaction \cite{li2018action,tu2019action,dhiman2020view}. Compared to traditional RGB or RGB-D video data, the skeleton sequences are compact motion data, which can dramatically reduce the computational cost of human action recognition. Besides, the skeleton data could show better performance when the background of human action is complicated \cite{yan2018spatial,xie2018memory}.

It is known that recurrent neural networks (RNN) \cite{liu2018multi,huang2018toward}, convolutional neural networks (CNN) \cite{nie2019view}, temporal convolutional networks (TCN)\cite{kim2017interpretable}, graph convolutional networks (GCN) \cite{yang2020sta}, and spatial temporal GCN (ST-GCN) \cite{yan2018spatial} have been applied to human action recognition. The earliest models for human action recognition often construct all coordinates of human joints from each frame for convolutional learning, like RNN or CNN. These learning models rarely investigate the endogenous relationship between joints and bones, leading to missing abundant unique and significant information in skeleton data. To understand the relationships between joints and bones, recent models build a skeleton graph, where nodes are human joints and edges are natural bones between two connected joints, and implement GCN to extract informative features. Later, the spatial-temporal GCN (ST-GCN) further developed GCN to simultaneously learn topological features of skeleton graphs \cite{yan2018spatial}. The ST-GCN constructed a skeleton graph to represent the connections of joints, and firstly proposed the temporal edges to link the relationships between consecutive frames. Although the ST-GCN captures the features of joints and bones, structurally key joints and bones are largely ignored, which may contain specific patterns and information of actions. For example, human hand joints and knee joints play key roles in the walking action. While the ST-GCN tries to capture all of the human joints, the correlations between key joints and actions might be weakened during the training process.

To address such issues, we need a new design that can automatically detect the key human joints and bones, and deeply emphasize their uncovered graph features in terms of their large distance for every human action. Meanwhile, the new model could encode the extracted graph features into training process as well as highlight their dynamic properties, which could fully consider different perspectives from skeleton graphs, and present a general graph embedding application. Additionally, while the skeletons are the structural graphs instead of 2D or 3D grids on different human actions, most of the previous models construct a constant graph as the input signal. Basically, to complete an action, everyone needs the cooperation of various body parts. The phenomenon indicates human actions are based on the relationships between different body parts, but some body parts may play key roles in human actions. For example, walking action strongly depends on legs, knees and feet, while sitting action is closely related to the head and trunk. An exogenous dependency has been proposed for disconnected joints or human parts, but they could not cover wider-range human actions \cite{zhang2019semantics}. These body parts could be considered as subgraphs of skeleton graphs.

Deep learning on graphs, in particular, have emerged as the
dominant models for learning representations on graphs \cite{kipf2016semi}. These graph learning models condense
neighborhood connectivity patterns into low-dimensional embedding features that can be used for a variety
of downstream tasks. While graph representation learning has made tremendous progress
in recent years \cite{wu2019simplifying}, prevailing models focus on learning useful features for nodes \cite{adhikari2018sub2vec}, edges \cite{vashishth2019composition}, subgraphs \cite{monti2018motifnet} or entire graphs \cite{pareja2020evolvegcn}. Graph-level features provide an overarching view of the graphs at macro details. In contrast, node-level features focus instead on the preservation of the local topological structure in micro details. Besides, subgraph features could effectively capture the unique topology of subgraphs, and incredibly improve the learning performance \cite{alsentzer2020subgraph}.

\begin{figure*}
\centering
\includegraphics[height=8.0cm,width=16cm]{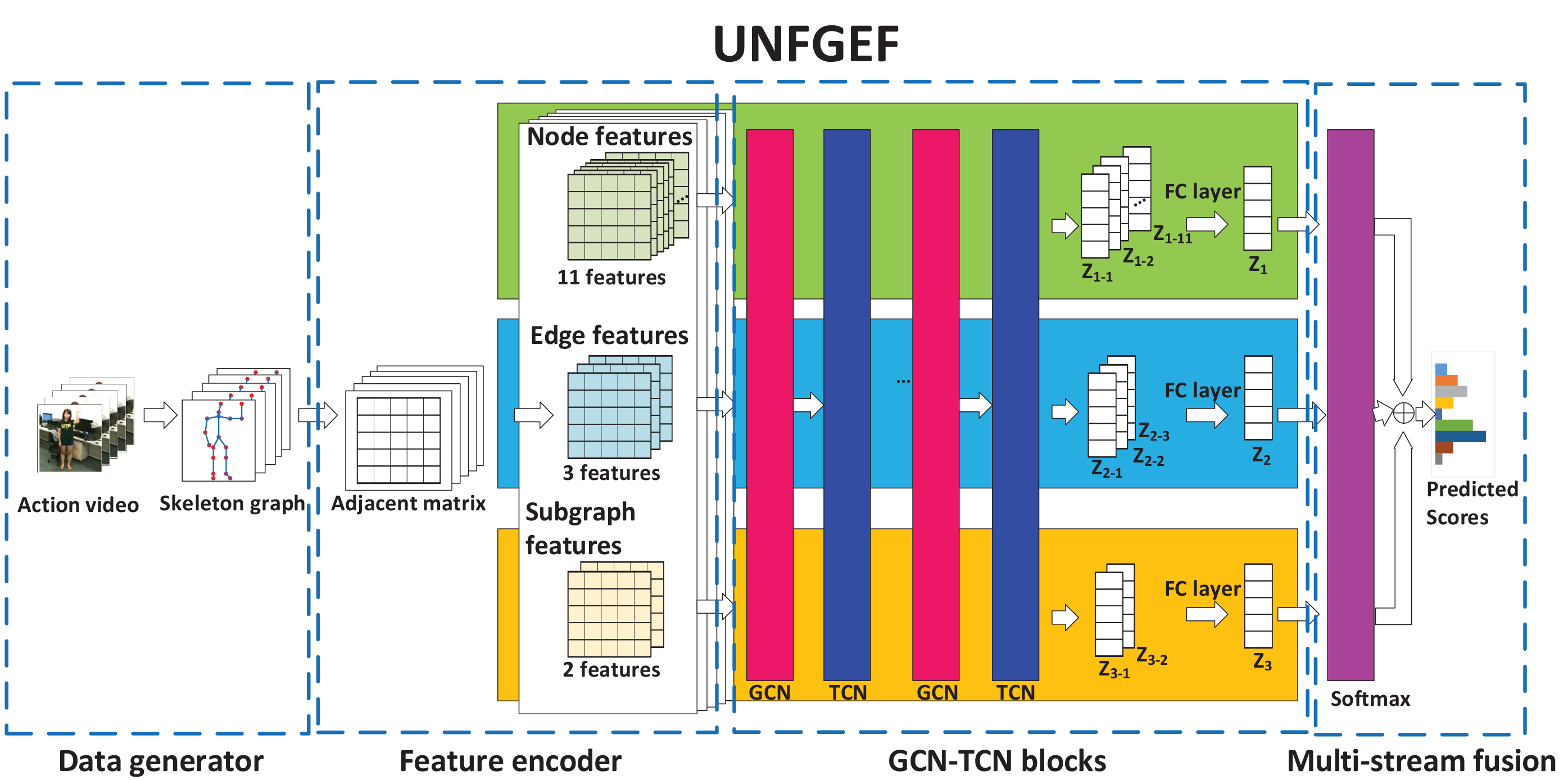}
\caption{The network architecture of UNFGEF model. The action video could be divided into each frame. Then, the related skeleton graph is built by each frame. We use adjacent matrix to represent the skeleton graph. Based on graph embedding algorithms, various graph features are extracted and categorized into three domains, such as node features, edge features and subgraph features. Next, these features are separately embedded into GCN and TCN networks. After the training process, the trained weighted matrices could be fully connected. Finally, the output is calculated by the Softmax classifier in each stream. And then the scores are added by the multi-stream fusion to make the final prediction.}
\label{fig1}
\end{figure*}

In this paper, we systematically study human actions, and propose a new framework to unify 15 graph embedding features with graph convolutional networks and to model characteristic skeletons for human action recognition, called UNFGEF. Compared to existing GCN models, this new model answers what characterizes node features, edge features and subgraph features for skeleton graphs. UNFGEF is expected to provide a ranking that identifies the most important joints, bones and body parts in different human actions. Inspired by graph embedding methods, we categorized the graph features into three domains, such as node centrality, edge centrality, subgraph features. The node centrality contains 10 node-level features, including Salton index, Jaccard index, and Katz index, etc \cite{fu2018link}. The edge centrality is calculated by the betweenness centrality, degree centrality, and similarity centrality. Finally, subgraph features are measured by clustering coefficient and motif index. Then, the extracted graph features are encoded into the graph convolution networks for further training. The UNFGEF can identify the relationships between physically connected and disconnected joints simultaneously, which effectively capture micro-level and high-level skeleton structures. Besides, this proposed module shows endogenous dependencies in an attention mechanism to enhance the learning performance of human action recognition. The pipeline of UNFGEF is illustrated in Fig.~\ref{fig1}.

The major contributions of this work are summarized as follows.

\begin{itemize}

    \item First, UNFGEF is the first work for embedding various graph features into graph evolutional networks. And it is designed to uncover the overlooked graph information between physically connected and disconnected parts of the human skeleton.
    \item Second, UNFGEF exploits 15 graph embedding methods in designing the embedding module to meet general demands in human actions. It could offer a new and deep direction of the action recognition task.
    \item Third, the motion information between consecutive frames is extracted for temporal information modeling. Both the spatial and temporal information are fed into a multi-stream framework for the action recognition task.
    \item Fourth, our model has state-of-the-art performance on three large-scale datasets for skeleton-based action recognition. The relative improvement is about 1\%. The finding indicates that these extracted graph features are fundamental factors in the skeleton data, which brings remarkable improvements for human action recognition.

\end{itemize}

Our results have implications for unifying graph embedding features with graph convolutional networks, advancing the science of human action recognition and improving our understanding of general graph learning.

\section{Related work}

\subsection{Skeleton-based action recognition}
Despite the illumination change and scene variation, the reliable skeleton data could be easily and accurately extracted by pose estimation algorithms from depth sensors \cite{kay2017kinetics} or RGB cameras \cite{Shahroudy_2016_NTURGBD,xiu2018pose} and temporal CNN \cite{li2017person}. The substantial skeleton data spurs the creation of new technological or theoretical action recognition models \cite{stergiou2020learning}. Recently, deep learning approaches have been widely applied to build the spatial-temporal frameworks of skeleton sequences in human action recognition \cite{dhiman2020view}. For example, RNN models are implemented into human action recognition due to effectively learning long-term sequence data \cite{li2017adaptive}. Compared to RNN, CNN also showed important implications for human action recognition, owing to its powerful parallelization over every element in the training process. Skeleton data are manually transferred into 2D or 3D images deployed by CNN \cite{yang2020sta}, which achieves high performance in action recognition. However, these images used by CNN cannot fully show the topological structures of skeleton data. To address the limitations, GCN successfully applies to human action recognition, and its plausible mechanisms lead to more promising results \cite{yan2018spatial}. The traditional GCN models only consider physical connections between different joints on human actions. Since the endogenous factors, referring to physically disconnected joints, also play a preeminent role in human action recognition, the skeleton structure conveys much unrevealed information and leaves potential space to improve the performance of GCN \cite{weng2017spatio}.

\subsection{Graph convolutional networks}
Mapping the relational data into graphs, the topological structures can be encoded to model the connections among nodes, and provide more promising perspectives underlying the data. Inspiring by this mechanism, GCN is successfully implemented in deep learning research.  The various approaches in modeling GCN fall into two categories, including spatial and spectral approaches. Spatial approaches use graph theory to define the nodes and edges for entities on data \cite{duvenaud2015convolutional,shi2020skeleton}. Interestingly, spectral approaches analyze the constructed graph in the frequency domain \cite{kipf2016semi}. The spectral approach usually leverages the Laplacian eigenvector to transform a graph in the time domain to in frequency domain, potentially resulting in large computation cost. Considering human action recognition, most of the methods choose the spatial approaches to construct the GCN due to the large size of skeleton data. Since the human body is naturally formed as a graph, not a sequence or an image, the related features are easily extracted from skeleton data. However, their works only focus on the static graph structure and are hard to understand the dynamic information and graph features from human actions.  Here, our model could adaptively extract the key information on joints, bones and body parts, which yields a dynamical GCN. Besides, the UNFGEF combines the high-order information to provide a new insight for learning human actions.

\subsection{Graph embedding methods}
From the perspective of extracted features, graph embedding models can be classified into two groups: topological embedding methods and content enhanced embedding approaches. First, topological embedding methods encode topological structure features from the graph, and the learning objective is to preserve the topological features. A number of probabilistic models have been designed, such as node2vec \cite{grover2016node2vec} and LINE \cite{tang2015line}. Recently deep learning models also implement the graph embedding features into their training process. These models investigate the first and second order graph features \cite{wang2016structural}, or reconstruct the positive node features \cite{cao2015grarep} via different variants of autoencoders. Meanwhile, community detection, motif extraction, and embedding models for subgraph have made tremendous progress in learning nodes \cite{adhikari2018sub2vec}, edges \cite{vashishth2019composition} or entire graphs \cite{monti2018motifnet}. These methods search for groups of nodes that are well-connected internally while being relatively well-separated from the rest of the graph and typically limited to individual connected components. Second, content enhanced embedding models simultaneously explore topological information and content features, which fully considers all available information during the training process. Henneberg growth model explores massive and complicated topological structures in large-scale graphs \cite{yang2018henneberg,lou2022controllability}. However, these graph embedding methods are hardly combined with graph convolutional networks for human action recognition. Our paper systematically explored available graph features of skeleton graphs, and encode them into graph convolutional architectures. The UNFGEF is different from the current learning representations of human action recognition.

\section{Graph embedding features}
In this paper, three groups of graph features are extracted from the skeleton data, such as node features, edge features and subgraph features. These graph features play a significant role in highlighting the key joints, bones and body parts in human actions. Besides, these graph features elaborately and systematically reflects all information hidden in the skeleton structure. These graph embedding methods have been proposed in many empirical studies on graph learning representatives \cite{fu2018link}.

A skeleton graph could be modeled by an undirected and weighted network $G(V,E)$, where $V$ is the number of joints, and $E$ is the number of bones, respectively. $e_{ij}$ is an edge between node $i$ and $j$. In the skeleton graph, each bone is linked by two joints, such as the source node and the target node, respectively. Note that every edge has the length of the corresponding bone information. For example, given two node (joints) coordinates, the source node $n_i=(x_i,y_i,z_i)$ and the target node $n_j=(x_j,y_j,z_j)$, the length of the edge (bone) can be calculated as $d(n_i,n_j)=\sqrt{(x_j-x_i)^2+(y_j-y_i)^2+(z_j-z_i)^2}$. In this skeleton graph, there is no cycle, which indicates every edge can only contain a source node and a target node. In this way, the graph adjacent matrix, node features, edge features and subgraph features are calculated, respectively.

\subsection{Node features}

A score $v_{ij}$ is calculated by the node feature algorithm. The score is to measure the common feature between two nodes. Next, the extracted features are listed below:

\begin{itemize}

    \item \textbf{Common neighbors}. It could be written as
        \begin{align}
            v_{ij}^{CN} = |\Gamma(i) \cap \Gamma(j)|,
        \label{eq1}
        \end{align}
        where $\Gamma(i)$ is the set of neighbors of node $i$ and $|\odot|$ is the cardinality of set. The feature matrix is set to $\hat{V}_1$.
    \item \textbf{Salton index}. It is defined as
        \begin{align}
            v_{ij}^{SA} = \frac{|\Gamma(i) \cap \Gamma(j)|}{\sqrt{k_i \times k_j}},
        \label{eq2}
        \end{align}
        where $k_i$ is the degree of $i$. Its matrix is set to $\hat{V}_2$.
    \item \textbf{Jaccard index}. It is defined as
        \begin{align}
            v_{ij}^{JI} = \frac{|\Gamma(i) \cap \Gamma(j)|}{|\Gamma(i) \cup \Gamma(j)|}.
        \label{eq3}
        \end{align}
        Its matrix is set to $\hat{V}_3$.
    \item \textbf{Hub promoted index}. It is defined as
        \begin{align}
            v_{ij}^{HPI} = \frac{|\Gamma(i) \cap \Gamma(j)|}{min(k_i,k_j)}.
        \label{eq4}
        \end{align}
        The calculated matrix is set to $\hat{V}_4$.
    \item \textbf{Srensen Index}. It is defined as
        \begin{align}
            v_{ij}^{SRI} = \frac{2|\Gamma(i) \cap \Gamma(j)|}{k_i+k_j}.
        \label{eq5}
        \end{align}
        Its matrix is set to $\hat{V}_5$.
    \item \textbf{Adamic-Adar Index}. It is defined as
        \begin{align}
            v_{ij}^{AA} = \sum_{z \in \Gamma(i) \cap \Gamma(j)}\frac{1}{log_{k_z}}.
        \label{eq6}
        \end{align}
        Its matrix is set to $\hat{V}_6$.
    \item \textbf{Local path index}. It is defined as
        \begin{align}
            v_{ij}^{LPI} = (A^2)_{ij}+\epsilon(A^3)_{ij},
        \label{eq7}
        \end{align}
        where $\epsilon$ is a hyper parameter and set to 0.1. $A$ is the adjacent matrix, and $(A^k)_{ij}$ is the number of path length $k$ linking with the nodes $i$ and $j$. Its matrix is set to $\hat{V}_7$.
        \item \textbf{Katz index}. It is defined as
        \begin{align}
            v_{ij}^{KI} = \alpha(A)_{ij}+\alpha^2(A^2)_{ij}+\alpha^3(A^3)_{ij}+\cdots,
        \label{eq8}
        \end{align}
        where $\alpha$ is a parameter. If $\alpha < \frac{1}{\lambda_{max}}$, it can be rewritten as
        \begin{align}
            v_{ij}^{KI} = (1-\alpha A)^{-1}-I_N,
        \label{eq9}
        \end{align}
        where $I_N$ is the identity matrix, and $\lambda_{max}$ is the largest eigenvalue of $A$. Its matrix is set to $\hat{V}_8$.
        \item \textbf{Preferential attachment index}. It is defined as
        \begin{align}
            v_{ij}^{PAI} = |\Gamma(i)| \cdot |\Gamma(j)|.
        \label{eq10}
        \end{align}
        Its matrix is set to $\hat{V}_9$.
        \item \textbf{Closeness centrality}. It is defined as
        \begin{align}
            v_{ij}^{cc} = \frac{N}{\sum_{j=1}^{N}d_{ij}},
        \label{eq11}
        \end{align}
        where $d_{ij}$ is the shortest path length between nodes $i$ and $j$. Its matrix is set to $\hat{V}_{10}$.

\end{itemize}

Based on 10 mentioned methods, we identify the node features from skeleton graphs. Next, we use attention mechanism to encode 10 node features (feature matrices) into GCN.

\subsection{Edge features}

In the skeleton graph, the bone can be served as an edge. Due to the different coordinates of each node, the distances of bones have a diverse length. Considering graph theory, the related algorithm is to identify the importance of edges in graph. Here, the edge features are calculated below:

\begin{itemize}

    \item \textbf{Betweenness centrality}. It is defined as
        \begin{align}
         e_{ij}^{BC} =\sum_{(l,q) \not= (i,j)}\frac{\widetilde{S}_{lq}(e_{ij}) d_{lq}(e_{ij})}{\widetilde{S}_{lq}d_{lq}},
        \label{eq12}
        \end{align}
        where $e_{ij}$ is the edge weight between node $i$ and node $j$, $\widetilde{S}_{lq}$ is the number of all existing shortest paths from node $l$ to node $q$, $\widetilde{S}_{lq}(e_{ij})$ is the number of all shortest paths from node $l$ to node $q$ that pass through edge $e_{ij}$, $d_{lq}$ is the total distance length from node $l$ to node $q$, and $d_{lq}(e_{ij})$ is the total distance length from node $l$ to node $q$ that pass through edge $e_{ij}$. Its matrix is set to $B_1$.
    \item \textbf{Degree centrality}. It is defined as
        \begin{align}
            e_{ij}^{DC} = \frac{e_{ij}}{\sum_{k}e_{ik}}.
        \label{eq13}
        \end{align}
        Its matrix is set to $B_2$.
    \item \textbf{Similarity centrality}. It is defined as
        \begin{align}
            e_{ij}^{SC} = \frac{f(i,j)e_{ij}}{\sum_{k}f(i,k)e_{ik}},
        \label{eq14}
        \end{align}
    where $f(i,j)$ is the similarity score function for node $i$ and $j$. Its matrix is set to $B_3$.
\end{itemize}

Three edge features could yield related weighted matrices. With frame changing, the weighted matrices automatically update. Additionally, these edge features play key roles in reflecting bone characteristics in human actions.

\subsection{Subgraph features}

Different body parts are associated with completing human actions. However, there exists a problem of how to evaluate the influence of every body part during human actions and measure the correlations between two body parts. In graph theory, the body part could be considered as a subgraph. A subgraph $S$ is a graph whose node set $N(S)$ is a subset of the node set $N(G)$, i.e. $N(S) \subseteq N(G)$. Similarly, the subgraph edge set $E(S)$ is a subset of the edge set $E(G)$, that is $E(S) \subseteq E(G)$. The subgraph features are extracted by the listing algorithms:

\begin{itemize}

    \item \textbf{Clustering coefficient}. It is defined as
        \begin{align}
          s_{i}^{CC}=\frac{2L_i}{k_i(k_i-1)},
        \label{eq15}
        \end{align}
        where $L_i$ is the number of edges between neighbors of node $i$. Its matrix is set to $S_1$.
    \item \textbf{Motif index}. It is defined as
        \begin{align}
          s_{i}^{MI}=M_{i}^{1}+M_{i}^{2}+M_{i}^{3}+\cdots,
        \label{eq16}
        \end{align}
        where $M_{i}^{k}$ denotes the frequency of node $i$ in subgraphs $M^k$, and $k$ is the number of nodes in subgraph. Its matrix is set to $S_2$.
\end{itemize}

The associations with subgraphs are measured by every node, that is, a weighted matrix with $N \times N$ dimensions.

\section{Method}

\subsection{Network architecture}

Shown in Fig.~\ref{fig1}, the pipeline of this network architecture contains four components, namely graph generator, feature encoder, GCN-TCN blocks, and multi-stream fusion. In the first component, the original action video would be transferred into the skeleton data. Then the adjacent matrix is generated from the skeleton data and is encoded into three streams for training and prediction. Second, the feature encoder calculated 15 different features mentioned in the above sections. These features are categorized into three domains, namely node features, edge features, and subgraph features. These extracted features would be trained by the GCN-TCN blocks, which could take the best advantages of the skeleton data to learn more information. The GCN-TCN blocks play key roles in training and updating parameters. Finally, the framework would fuse the output scores of the softmax layers from the three streams. In the following sections, the last two components of the proposed network architecture are elaborated in detail.

\subsection{GCN-TCN blocks}

\begin{figure*}
\centering
\includegraphics[height=6.6cm,width=17cm]{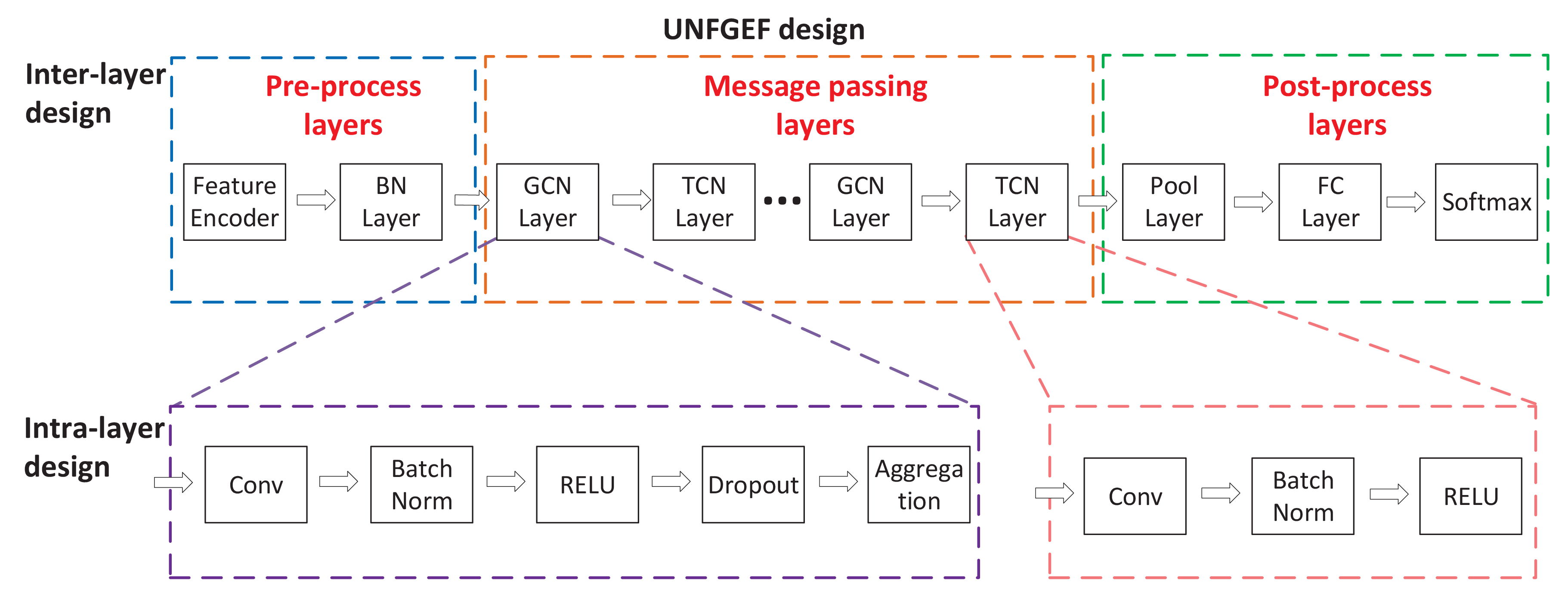}
\caption{The layer design of UNFGEF. It contains two types of layers, such as inter-layer design and intra-layer design. In inter-layer design, there are three steps, including pre-processing layers, message passing layers and post-processing layers. The core layer is the message passing layer, consisting of multiple GCN-TCN blocks. For intra-layer design, it shows the details of GCN and TCN layers. Compared to TCN layer, GCN layer contains dropout and aggregation layers. }
\label{fig2}
\end{figure*}

One basic GCN-TCN block is composed of two layers, namely the GCN layer and TCN layer, based on ST-GCN model \cite{yan2018spatial}. Its pipeline is illustrated in Fig.~\ref{fig2}. Compared to the original GCN layer, the GCN layer has 5 layers, and appends two more layers, including the dropout layer and aggregation layer. The next TCN layer has 3 layers, namely convolution, batch normalization, and RELU layers. Each GCN-TCN block adds a residual connection to avoid gradient vanishing.

Let $x \in R^N$ be a feature vector for every node in a graph. The spectral convolution on a graph can be formulated as
\begin{align}
 Z&=\hat{D}^{-\frac{1}{2}}\hat{A}\hat{D}^{-\frac{1}{2}}X\Theta \nonumber \\
    &=\widetilde{A}X\Theta,
\label{eq21}
\end{align}
where $\widetilde{A}=\hat{D}^{-\frac{1}{2}}\hat{A}\hat{D}^{-\frac{1}{2}}=I_{N \times N}+D^{-\frac{1}{2}}AD^{-\frac{1}{2}}$ ($A$ is considered as its adjacency matrix with $N \times N$ dimensions), $\hat{D_i}=\sum_j \hat{A}_{ij}$, $X \in R^{N \times C}$ and $\Theta \in R^{C \times F}$ ($C$ is the dimensions of feature vector per node and $F$ is the filter channels). Note that $Z$ is the convolved or extracted feature matrix, and $\Theta$ is the filtering-parameters matrix.  Thus, the complexity of this normalized formulation is $O(EFC)$.

To implement the UNFGEF in human action recognition, let $X=X_{in}$ be the input feature matrix on total frames per one video sample. For encoding graph embedding features, $\widetilde{A}$ of the Eq.~\ref{eq21} would be modified. For example, the node features could be encoded by this method
\begin{align}
 Z_1&=(\widetilde{A}+J)X_{in}\Theta_1,
\label{eq22}
\end{align}
where $J=\sum_k \hat{V}_k$ is 10 node features, and $\hat{V}_k$ means the extracted node feature matrix, like Salton index and Jaccard index, etc. $Z_1$ and $\Theta_1$ are the output and parameters of the node stream, respectively.

Similar to node features, let $B$ be 3 edge features and $S$ be 2 subgraph features. The Eq.~\ref{eq22} could be rewritten as
\begin{align}
 Z_2&=(\widetilde{A}+B)X_{in}\Theta_2,
 \label{eq23}
\end{align}
\begin{align}
 Z_3&=(\widetilde{A}+S)X_{in}\Theta_3,
 \label{eq24}
\end{align}
where $B=\sum_{k}B_k$, $S=\sum_{k}S_k$, and $B_k$ and $S_k$ represent the extracted edge and subgraph feature matrices, respectively. $Z_2$ and $\Theta_2$ are the output and parameters of the edge stream, whereas $Z_3$ and $\Theta_3$ are the output and parameters of the subgraph stream.

\subsection{Multi-stream Fusion}

Compared to previous works, the three streams, such as node, edge, and subgraph streams, consist of 15 graph features, which is the first to systematically investigate the skeleton structures and uncover neglected information. The three streams are calculated from mentioned graph embedding features. Then, 15 graph embedding features pass about 10 GCN-TCN blocks. Each softmax score is calculated by
\begin{align}
 P_1 &=Softmax \left (ReLU(Z_1)\right), \\
 P_2 &=Softmax \left (ReLU(Z_2)\right), \\
 P_3 &=Softmax \left (ReLU(Z_3)\right),
 \label{eq25}
\end{align}
where $Z_1$, $Z_2$, $Z_3$ are outputs of node, edge, and subgraph streams, respectively. The multi-class cross-entropy loss is written as
\begin{align}
 L &=-\sum_{i \in y_L} Y_ilog(p_i),
 \label{eq26}
\end{align}
where $p_i$ is the predicted probability of class $i$ and $y_L$ is the number of action classes. if $Y = i$ , $Y_i$ is set as 1, otherwise it is 0.

Fused by three softmax scores, the final score can be obtained
\begin{align}
 R &=P_1+P_2+P_3,
 \label{eq27}
\end{align}
where $P_1$, $P_2$, $P_3$ are softmax scores of node, edge, and subgraph streams, respectively. The final score could yield the score ranking and predict the action classification.

\section{Experiment}
To evaluate the proposed framework, i.e. UNFGEF, we conduct the experiments on three large-scale action recognition datasets: NTU-RGB+D \cite{Shahroudy_2016_NTURGBD}, Kinetics \cite{kay2017kinetics} and SYSU-3D \cite{hu2015jointly}. First, since the size of the NTU-RGB+D dataset is smaller than that of the Kinetics dataset, we perform the ablation studies of our model on the NTU-RGB+D dataset, which examines every component of the proposed model on the action recognition performance. Then, the performance of UNFGEF is verified and compared with that of state-of-the-art approaches.

\subsection{Datasets}
\subsubsection{NTU-RGB+D} NTU-RGB+D is currently one of the largest dataset widely used in the skeleton-based action recognition, which consists of 56,880 video samples categorized into 60 action classes. All of video samples are acted by 40 volunteers, who come from different age groups from 10 to 35. Each action is recorded by three Kinect V2 cameras concurrently from different horizontal angles, like $-45^{\circ},0^{\circ},45^{\circ}$. This dataset contains 3D skeletal data for each video sample. Here, it records 25 joints for each frame on each subject sample, while each video sample has one or two subjects. The dataset recommends two benchmarks. The first one is the cross-subject (X-sub), which is composed of a training set (40,320 video samples) and a testing set (16,560 video samples). The second one is the cross-view (X-view), where the training set contains 37,920 video samples captured by cameras No. 2 and No. 3, and the testing set contains 18,960 video samples captured by camera No. 1. In the comparison, we report the top-1 accuracy on two benchmarks.

\subsubsection{Kinetics} Kinetics is another large-scale dataset for human action recognition, containing 300,000 video samples. All of video samples cover 400 human action classes. The video samples are recorded by YouTube and have various subjects. It only contains RGB video samples without raw skeleton data. We obtain 18 joints' positions of every frame using the publicly available OpenPose toolbox. The captured skeleton data contains two dimensions of coordinates, i.e., ($x,y$), and the confidence score ($c$) for each joint. The dataset consists of a training set (240,000 video samples) and a testing set (20,000 video samples). Following the evaluation method, we report the top-1 and top-5 accuracy on the benchmark.

\subsubsection{SYSU-3D} SYSU-3D is a dataset recorded by Kinetic camera. It consists of 480 skeleton video samples of 12 action categories performed by 40 subjects. Each video sample has 20 joints. There are two standard evaluation cases for this dataset, i.e., cross-subject (CS) setting and same-subject (SS) setting. For the cross-subject setting, half of the subjects are used for training and the rest are for testing. For the same subject setting, half of the samples from each activity are used for training and the rest are for testing.

\subsection{Training details}
UNFGEF is implemented on the PyTorch deep learning framework. The kernel size of this model is set to 4. And it contains three streams to train the skeleton data. The optimization method uses the stochastic gradient descent with Nesterov momentum (0.9). And the loss function chooses the cross-entropy function for back propagating gradients. The batch size is 32 or 64. The learning rate is also set to 0.1 and the decay rate is 0.0001. The UNFGEF model is trained on 2 GTX-1080Ti GPUs. For different datasets, we set the specific configurations of UNFGEF.

The number of GCN-TCN block are 10. The numbers of output channels are 64, 64, 64, 64, 128, 128, 128, 256, 256 and 256 in each block. Note that the drop rate of GCN layers is set to 0.4. At the beginning of 10 basic blocks, there exists a batch normalization layer, which normalizes the input skeleton data. Through 10 basic blocks, it is followed by a global average pooling layer, adjusting different feature dimensions of video samples into the same dimensions. Finally, the final result is calculated by the Softmax layer to yield the classification results.

For the NTU-RGB+D dataset, each video sample contains no more than two persons. The feature dimensions are 64, 128 and 256, respectively. The number of frames in each video sample is set to 300. If the frame number is less than 300 frames, we repeat the video sample until it reaches 300 frames. In Eq.~\ref{eq22}, $X_{in}$ should be the total features from 300 frames. The training process ends at the 50th epoch.

For the Kinetics dataset, the input configuration of Kinetics is set to 150 frames with 2 persons in each video sample. Here, we randomly select 150 frames from the input skeleton data, and disorder the joint coordinates with randomly chosen translations and rotations. Here, $X_{in}$ should be the total features from 150 frames in Eq.~\ref{eq22}. The training process converges to the 60th epoch.

For the SYSU-3D dataset, it contains one person per each frame, and has two standard evaluation protocols, i.e., Setting-1 (cross-subject) and Setting-2 (same-subject). We randomly select 200 frames from the input skeleton data, and disorder the joint coordinates with randomly chosen translations and rotations. The training process stops at the 60th epoch.

\subsection{Ablation study}

We examine the influence of each graph embedding feature in UNFGEF with the X-view benchmark on the NTU-RGB+D dataset. The performance of ST-GCN on the NTU-RGB+D dataset is 88.3\%. By highlighting 15 graph embedding features with multi-stream framework, the result is improved to 96.4\%, which brings a definite improvement. The detail is introduced in the below parts.

\subsubsection{Graph embedding features}

As mentioned in Section 3, there are three types of graph embedding features in the UNFGEF framework, i.e., node features $J$, edge features $B$, and subgraph features $S$. Here, we only investigate effects of these graph features. The different combinations of the graph features are tested on the UNFGEF framework, including UNFGEF only with $J$, UNFGEF only with $B$, UNFGEF only with $S$, UNFGEF with $J$ and $B$, UNFGEF with $J$ and $S$, UNFGEF with $B$ and $S$, and UNFGEF with $J$, $B$ and $S$. And the results are shown in Table~\ref{table1}. UNFGEF ($J$+$B$+$S$) outperforms the baseline model ST-GCN by 8.2\% on the X-view NTU-RGB+D dataset, which strongly demonstrate the importance of graph embedding features. The results show that each graph feature extracted from the graph is significant for human action recognition. Besides, deleting any one of streams will dramatically reduce the performance.

\begin{figure}[!h]
\centering

\includegraphics[width=9cm]{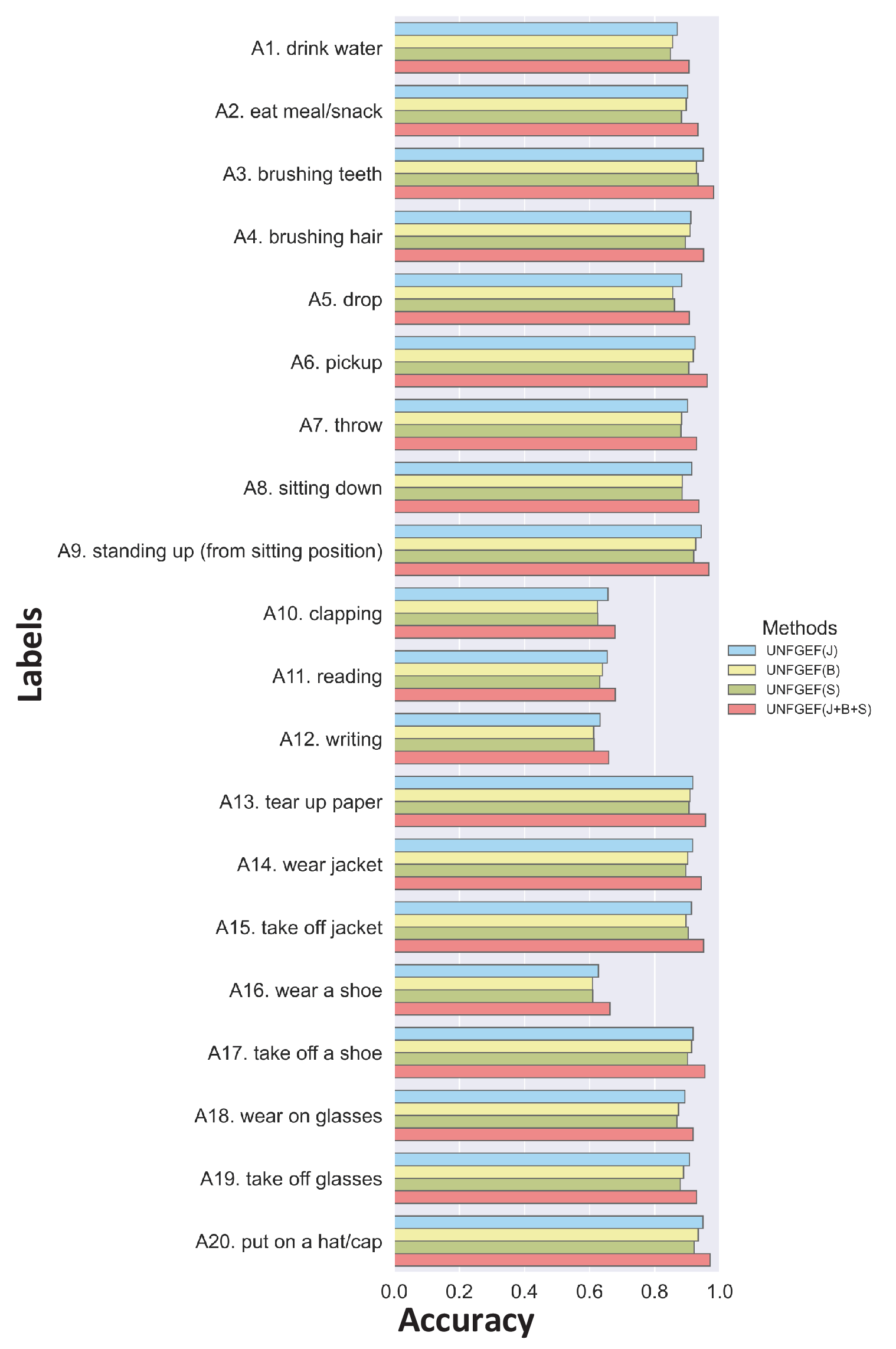}
\caption{The ablation study of four methods on 20 action classes. The four methods are UNFGEF only with $J$, UNFGEF only with $B$, UNFGEF only with $S$, and UNFGEF with $J$, $B$ and $S$. The x axis is the top-1 accuracy on the X-view NTU-RGB+D dataset.}
\label{fig3}
\end{figure}

\begin{figure*}
\centering
\includegraphics[width=17.5cm]{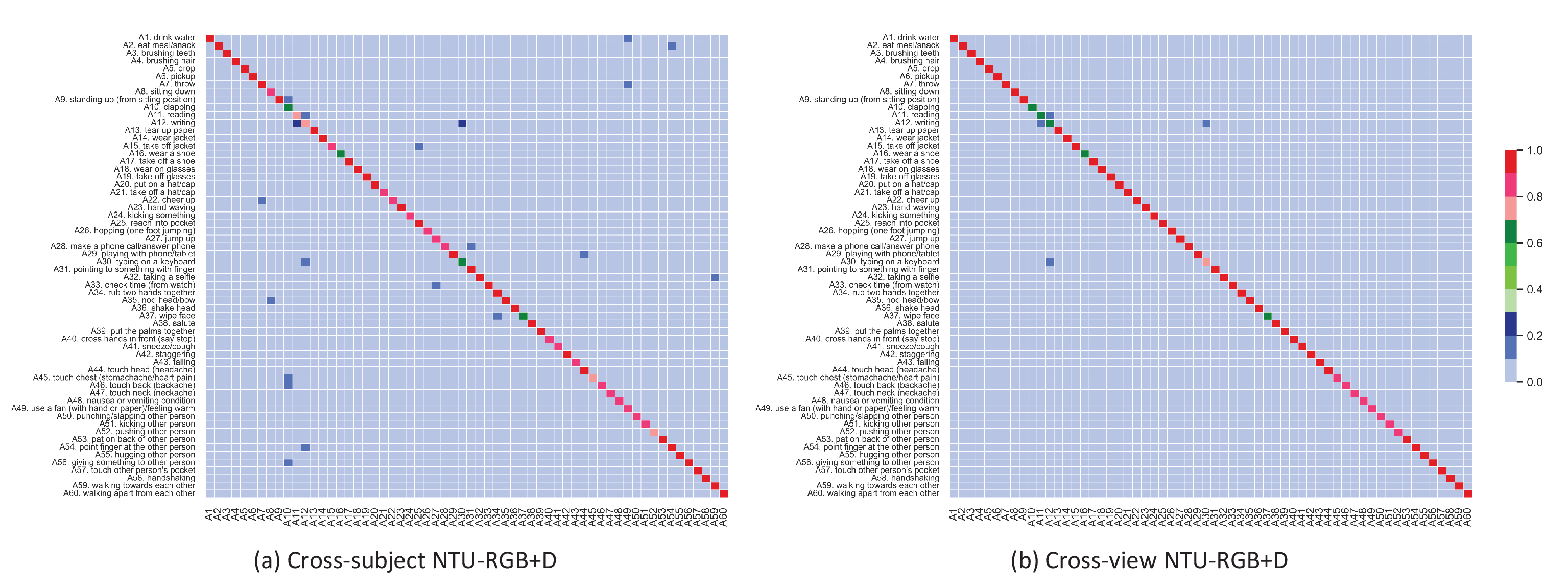}
\caption{Confusion matrices of the UNFGEF framework on the NTU RGB+D dataset. We can see that the most significant classification errors occur among classes that are physically similar. (a) Cross-subject evaluation. (b) Cross-view evaluation.}
\label{fig4}
\end{figure*}

Shown in fig.~\ref{fig3}, the accuracy of our new framework achieves more than 93.0\% on the majority of action classes, except A.10 to A.12 and A.16. Most importantly, the node features have a greater impact than edge features and subgraph features in the UNFGEF framework. If the UNFGEF framework only has subgraph features, the performance is not good. After combining all graph features together, we can utilize all in information and achieve better outcomes. The result indicates that the learned graph features are important, which also demonstrates the significance of the node features. The confusion matrices of UNFGEF on 60 action classes of the NTU-RGB+D dataset
are shown in Fig.~\ref{fig4}. The performance on the cross-view benchmark is better than that on the cross-subject benchmark.
\begin{table}[!tbp]
\begin{center}
\caption{Comparisons of the top-1 accuracy on the X-view NTU-RGB+D dataset when the different combinations of graph embedding features are tested on UNFGEF.}
\label{table1}
\begin{tabular}{ll}
\hline\noalign{\smallskip}
Methods & Accuracy(\%)\\
\noalign{\smallskip}
\hline
\noalign{\smallskip}
ST-GCN & 88.3\\
UNFGEF($J$) &  95.7\\
UNFGEF($B$) & 94.9 \\
UNFGEF($S$) & 94.1\\
UNFGEF($J$+$B$) &  95.9\\
UNFGEF($J$+$S$) & 95.7 \\
UNFGEF($B$+$S$) & 94.9\\
\pmb{UNFGEF($J$+$B$+$S$)}  &  \pmb{96.5}\\
\hline
\end{tabular}
\end{center}
\end{table}

\subsubsection{Visualization of the centrality graphs.}

\begin{figure*}
\centering
\includegraphics[height=10.13cm,width=14cm]{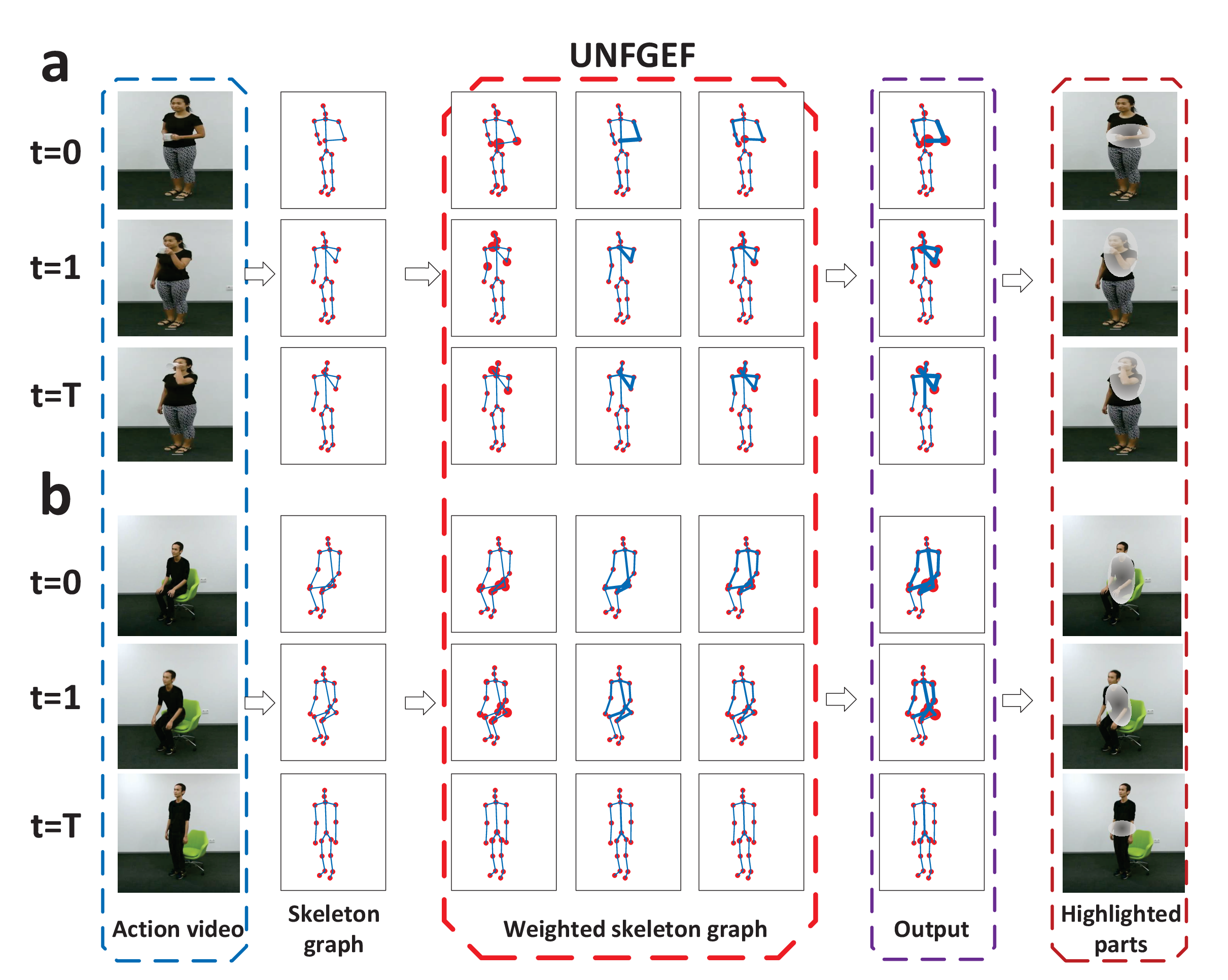}
\caption{Visualization of weighted skeleton graphs during training process on the NTU RGB+D dataset. The size of nodes and edges indicates its parameter weight. From left to right, skeleton graphs reflect the node features, edge features, and subgraph features, respectively. Each sequence visualizes an example of an action in NTU RGB+D. (a) Drink water. (b) Standing up (from sitting position). Three frames are respectively selected at $t=0,t=1,t=T$. Each skeleton graph contains 18 key joints.}
\label{fig5}
\end{figure*}

Fig.~\ref{fig5} is a visualization of three types of graph embedding features for one layer on the same action, i.e. node features, edge features, and subgraph features, respectively. The skeleton graphs are plotted from the physical connections of the human body. The red joints, bones, and body parts represent the highlighted parts by these graph features. In the drinking-water action class, the highlights of body parts move from the left arm to the upper body. For the standing-up action, the low trunk and upper legs play the key roles during the action. It indicates that the node and edge of the human trunk play significant roles in this action. Besides, it also suggests that a traditional skeleton graph is not the best choice for the action recognition task, and different actions need graphs with different skeleton structures. The node features pay more attention to the adjacent joints in the physical skeleton graph. For the subgraph features, the neck part and the hipbone part are detected to have a stronger connection, although they are far away from each other in the physical skeleton. Compared to previous works, these extracted graph features clearly capture the low-level features, like key joints and bones, and high-level structures, namely body parts. This information could uncover endogenous factors in human actions and give new perspectives on human action recognition and graph convolutional networks. Thus, graph embedding features are more relevant to the action classification task.

\subsection{Comparison with the state-of-the-art}

\begin{table}[!h]
\begin{center}
\caption{Comparisons of the top-1 accuracy with state-of-the-art methods on the NTU-RGB+D dataset.}
\label{table2}
\begin{tabular}{lll}
\hline\noalign{\smallskip}
Methods & X-sub (\%)& X-view (\%)\\
\noalign{\smallskip}
\hline
\noalign{\smallskip}
Deep LSTM \cite{Shahroudy_2016_NTURGBD} & 60.7  &67.3\\
2s-3DCNN \cite{liu2017two} &66.8  &72.6\\
ST-LSTM \cite{liu2016spatio} & 69.2  &77.7\\
STA-LSTM \cite{song2017end} &  73.4  &81.2\\
VA-LSTM \cite{zhang2017view} &  79.2  &87.7\\
ARRN-LSTM \cite{li2018skeleton} & 80.7 & 88.8\\
Ind-RNN \cite{li2018independently} &  81.8  &88.0\\
TCN \cite{kim2017interpretable}  &74.3 & 83.1\\
C-CNN+MTLN \cite{ke2017new}  & 79.6 & 84.8\\
Synthesized CNN \cite{liu2017enhanced}  &80.0  &87.2\\
ST-GCN \cite{yan2018spatial}  & 81.5  &88.3\\
CNN-Motion+Trans \cite{du2015skeleton} & 83.2 & 89.3\\
ResNet152 \cite{li2017skeleton}  &85.0 & 92.3\\
DPRL+GCNN \cite{tang2018deep}  & 83.5  &89.8\\
2s-AGCN \cite{shi2019two} &88.5&95.1\\
\hline
\pmb{UNFGEF (ours)} & \pmb{90.4}  &\pmb{96.5}\\
\hline
\end{tabular}
\end{center}
\end{table}

\begin{table}[!h]
\begin{center}
\caption{Comparisons of the top-1 and top-5 accuracy with state-of-the-art methods on the Kinetics-Skeleton dataset.}
\label{table3}
\begin{tabular}{lll}
\hline\noalign{\smallskip}
Methods & Top-1 (\%) & Top-5 (\%) \\
\noalign{\smallskip}
\hline
\noalign{\smallskip}
Feature Encoding \cite{fernando2015modeling}& 14.9 &25.8\\
Deep LSTM \cite{Shahroudy_2016_NTURGBD}& 16.4 &35.3\\
TCN \cite{kim2017interpretable} & 20.3 &40.0\\
ST-GCN \cite{yan2018spatial}  & 30.7& 52.8\\
AS-GCN \cite{Li_2019_CVPR} &34.8 &56.3\\
2s-AGCN \cite{shi2019two} &36.1& 58.7\\
\hline
\pmb{UNFGEF (ours)} & \pmb{37.6}  &\pmb{60.5}\\
\hline
\end{tabular}
\end{center}
\end{table}

\begin{table}[!h]
\begin{center}
\caption{Comparisons of state-of-the-art methods on the SYSU-3D dataset.}
\label{table4}
\begin{tabular}{lll}
\hline\noalign{\smallskip}
Methods & Setting-1 (\%)& Setting-2 (\%)\\
\noalign{\smallskip}
\hline
\noalign{\smallskip}
VA-RNN(aug.) \cite{zhang2019view} & 80.5  &79.7\\
VA-CNN(aug.) \cite{zhang2019view} &  85.1  &84.0\\
EleAttG \cite{zhang2018adding} &  85.7  &85.7\\
SGN \cite{zhang2020semantics}  & 86.9  &86.5\\
SLnL-rFA+ML \cite{hu2019joint} &88.3&88.1\\
\hline
\pmb{UNFGEF (ours)} & \pmb{92.1}  &\pmb{91.4}\\
\hline
\end{tabular}
\end{center}
\end{table}

We compare the performance of our model with state-of-the-art action recognition methods on the
NTU-RGB+D dataset \cite{Shahroudy_2016_NTURGBD,liu2017two,liu2016spatio,song2017end,zhang2017view,li2018skeleton,li2018independently,kim2017interpretable,ke2017new,liu2017enhanced,yan2018spatial,du2015skeleton,li2017skeleton,tang2018deep,shi2019two} , Kinetics dataset \cite{fernando2015modeling,Shahroudy_2016_NTURGBD,kim2017interpretable,yan2018spatial,Li_2019_CVPR,shi2019two}
, and SYSU dataset \cite{zhang2019view,zhang2018adding,zhang2020semantics,hu2019joint}. The results of three comparisons are shown in Table~\ref{table2}, Table~\ref{table3} and Table~\ref{table4}, respectively. The methods used for comparison are categorized as RNN-based methods \cite{Shahroudy_2016_NTURGBD, liu2016spatio,song2017end,zhang2017view,li2018skeleton,fernando2015modeling,zhang2019view,zhang2018adding}, CNN-based methods \cite{liu2017two,li2018independently,kim2017interpretable,ke2017new,liu2017enhanced,du2015skeleton,li2017skeleton,tang2018deep}, and GCN-based methods \cite{yan2018spatial,shi2019two,Li_2019_CVPR,zhang2020semantics,hu2019joint}. Our model outperforms state-of-the-art methods with a large margin on these datasets, which validates the superiority of our model. For the NTU-RGB+D dataset, the proposed UNFGEF framework achieves 8.9\% and 8.2\% improvements over the most influence method, i.e. ST-GCN, on the cross-subject and cross-view benchmarks, respectively. In \cite{shi2019two}, the authors encode joint and bone features into GCN and exploit two-stream framework for skeleton-based human action recognition. Compared to this model, our model outperforms it by 1.9\% and 1.4\% on both benchmarks.

In Table~\ref{table3}, UNFGEF surpasses recent GCN-based methods, such as ST-GCN \cite{yan2018spatial}, 2s-AGCN \cite{shi2019two}, and AS-GCN \cite{Li_2019_CVPR}. Compared to 2s-AGCN \cite{shi2019two}, the accuracy of the proposed model relatively increases by 1.5\% and 1.8\% on Top-1 and Top-5 benchmarks. Our model also achieves state-of-the-art performance on the Kinetics-Skeleton dataset. On the SYSU dataset, the recognition results are listed in Table~\ref{table4}. Even though SLnL-rFA+ML \cite{hu2019joint} achieves to 88.3\% and 88.1\% on setting-1 and setting-2 benchmarks, the UNFGEF model raises by 3.8\% and 3.4\%, respectively.

\section{Conclusions}

In this work, we proposed a novel framework of unifying graph embedding features (UNFGEF) with graph convolutional networks for human action recognition. UNFGEF captures the key joints, bones, and body parts from human actions and embeds them into graph convolution networks to adaptively learn fusion features. On the contrary, the existing methods always ignore the importance of graph information on skeleton data, i.e., node features, bone features, and subgraph features. However, these embedding features could show the low-level features and high-level information of the skeleton data. Our UNFGEF uses a multi-stream framework to explicitly exploit 15 graph embedding structures, which improves the performance significantly. UNFGEF is validated on three large-scale datasets, namely NTU-RGB+D, Kinetics, and SYSU-3D, and outperforms state-of-the-art methods on all the three datasets. The finding indicates that these extracted graph features are fundamental mechanisms and factors hidden in the skeleton topology. This novel framework provides some new insights on future studies of skeleton-based action recognition. For example, how to incorporate the correlations between adjacent frames, such as similarities, distances, and structures into the UNFGEF framework becomes a natural question.


\ifCLASSOPTIONcaptionsoff
  \newpage
\fi



%

\bibliographystyle{IEEEtran}
\bibliography{IEEEabrv,egbib}

\begin{thebibliography}{10}
\providecommand{\url}[1]{#1}
\csname url@samestyle\endcsname
\providecommand{\newblock}{\relax}
\providecommand{\bibinfo}[2]{#2}
\providecommand{\BIBentrySTDinterwordspacing}{\spaceskip=0pt\relax}
\providecommand{\BIBentryALTinterwordstretchfactor}{4}
\providecommand{\BIBentryALTinterwordspacing}{\spaceskip=\fontdimen2\font plus
\BIBentryALTinterwordstretchfactor\fontdimen3\font minus
  \fontdimen4\font\relax}
\providecommand{\BIBforeignlanguage}[2]{{%
\expandafter\ifx\csname l@#1\endcsname\relax
\typeout{** WARNING: IEEEtran.bst: No hyphenation pattern has been}%
\typeout{** loaded for the language `#1'. Using the pattern for}%
\typeout{** the default language instead.}%
\else
\language=\csname l@#1\endcsname
\fi
#2}}
\providecommand{\BIBdecl}{\relax}
\BIBdecl

\bibitem{yan2018spatial}
S.~Yan, Y.~Xiong, and D.~Lin, ``Spatial temporal graph convolutional networks
  for skeleton-based action recognition,'' in \emph{Thirty-second AAAI
  conference on artificial intelligence}, 2018.

\bibitem{luo2019tangent}
G.~Luo, J.~Wei, W.~Hu, and S.~J. Maybank, ``Tangent fisher vector on matrix
  manifolds for action recognition,'' \emph{IEEE Transactions on Image
  Processing}, vol.~29, pp. 3052--3064, 2019.

\bibitem{song2017end}
S.~Song, C.~Lan, J.~Xing, W.~Zeng, and J.~Liu, ``An end-to-end spatio-temporal
  attention model for human action recognition from skeleton data,'' in
  \emph{Thirty-first AAAI conference on artificial intelligence}, 2017.

\bibitem{li2018action}
C.~Li, Z.~Cui, W.~Zheng, C.~Xu, R.~Ji, and J.~Yang, ``Action-attending graphic
  neural network,'' \emph{IEEE Transactions on Image Processing}, vol.~27,
  no.~7, pp. 3657--3670, 2018.

\bibitem{tu2019action}
Z.~Tu, H.~Li, D.~Zhang, J.~Dauwels, B.~Li, and J.~Yuan, ``Action-stage
  emphasized spatiotemporal vlad for video action recognition,'' \emph{IEEE
  Transactions on Image Processing}, vol.~28, no.~6, pp. 2799--2812, 2019.

\bibitem{dhiman2020view}
C.~Dhiman and D.~K. Vishwakarma, ``View-invariant deep architecture for human
  action recognition using two-stream motion and shape temporal dynamics,''
  \emph{IEEE Transactions on Image Processing}, vol.~29, pp. 3835--3844, 2020.

\bibitem{xie2018memory}
C.~Xie, C.~Li, B.~Zhang, C.~Chen, J.~Han, C.~Zou, and J.~Liu, ``Memory
  attention networks for skeleton-based action recognition,'' \emph{arXiv
  preprint arXiv:1804.08254}, 2018.

\bibitem{liu2018multi}
A.-A. Liu, N.~Xu, W.-Z. Nie, Y.-T. Su, and Y.-D. Zhang, ``Multi-domain and
  multi-task learning for human action recognition,'' \emph{IEEE Transactions
  on Image Processing}, vol.~28, no.~2, pp. 853--867, 2018.

\bibitem{huang2018toward}
W.~Huang, L.~Fan, M.~Harandi, L.~Ma, H.~Liu, W.~Liu, and C.~Gan, ``Toward
  efficient action recognition: Principal backpropagation for training
  two-stream networks,'' \emph{IEEE Transactions on Image Processing}, vol.~28,
  no.~4, pp. 1773--1782, 2018.

\bibitem{nie2019view}
Q.~Nie, J.~Wang, X.~Wang, and Y.~Liu, ``View-invariant human action recognition
  based on a 3d bio-constrained skeleton model,'' \emph{IEEE Transactions on
  Image Processing}, vol.~28, no.~8, pp. 3959--3972, 2019.

\bibitem{kim2017interpretable}
T.~S. Kim and A.~Reiter, ``Interpretable 3d human action analysis with temporal
  convolutional networks,'' in \emph{2017 IEEE conference on computer vision
  and pattern recognition workshops (CVPRW)}.\hskip 1em plus 0.5em minus
  0.4em\relax IEEE, 2017, pp. 1623--1631.

\bibitem{yang2020sta}
H.~Yang, C.~Yuan, L.~Zhang, Y.~Sun, W.~Hu, and S.~J. Maybank, ``Sta-cnn:
  Convolutional spatial-temporal attention learning for action recognition,''
  \emph{IEEE Transactions on Image Processing}, vol.~29, pp. 5783--5793, 2020.

\bibitem{zhang2019semantics}
P.~Zhang, C.~Lan, W.~Zeng, J.~Xue, and N.~Zheng, ``Semantics-guided neural
  networks for efficient skeleton-based human action recognition,'' \emph{arXiv
  preprint arXiv:1904.01189}, 2019.

\bibitem{kipf2016semi}
T.~N. Kipf and M.~Welling, ``Semi-supervised classification with graph
  convolutional networks,'' \emph{arXiv preprint arXiv:1609.02907}, 2016.

\bibitem{wu2019simplifying}
F.~Wu, A.~Souza, T.~Zhang, C.~Fifty, T.~Yu, and K.~Weinberger, ``Simplifying
  graph convolutional networks,'' in \emph{International conference on machine
  learning}.\hskip 1em plus 0.5em minus 0.4em\relax PMLR, 2019, pp. 6861--6871.

\bibitem{adhikari2018sub2vec}
B.~Adhikari, Y.~Zhang, N.~Ramakrishnan, and B.~A. Prakash, ``Sub2vec: Feature
  learning for subgraphs,'' in \emph{Pacific-Asia Conference on Knowledge
  Discovery and Data Mining}.\hskip 1em plus 0.5em minus 0.4em\relax Springer,
  2018, pp. 170--182.

\bibitem{vashishth2019composition}
S.~Vashishth, S.~Sanyal, V.~Nitin, and P.~Talukdar, ``Composition-based
  multi-relational graph convolutional networks,'' \emph{arXiv preprint
  arXiv:1911.03082}, 2019.

\bibitem{monti2018motifnet}
F.~Monti, K.~Otness, and M.~M. Bronstein, ``Motifnet: a motif-based graph
  convolutional network for directed graphs,'' in \emph{2018 IEEE Data Science
  Workshop (DSW)}.\hskip 1em plus 0.5em minus 0.4em\relax IEEE, 2018, pp.
  225--228.

\bibitem{pareja2020evolvegcn}
A.~Pareja, G.~Domeniconi, J.~Chen, T.~Ma, T.~Suzumura, H.~Kanezashi, T.~Kaler,
  T.~Schardl, and C.~Leiserson, ``Evolvegcn: Evolving graph convolutional
  networks for dynamic graphs,'' in \emph{Proceedings of the AAAI Conference on
  Artificial Intelligence}, vol.~34, no.~04, 2020, pp. 5363--5370.

\bibitem{alsentzer2020subgraph}
E.~Alsentzer, S.~Finlayson, M.~Li, and M.~Zitnik, ``Subgraph neural networks,''
  \emph{Advances in Neural Information Processing Systems}, vol.~33, pp.
  8017--8029, 2020.

\bibitem{fu2018link}
C.~Fu, M.~Zhao, L.~Fan, X.~Chen, J.~Chen, Z.~Wu, Y.~Xia, and Q.~Xuan, ``Link
  weight prediction using supervised learning methods and its application to
  yelp layered network,'' \emph{IEEE Transactions on Knowledge and Data
  Engineering}, vol.~30, no.~8, pp. 1507--1518, 2018.

\bibitem{kay2017kinetics}
W.~Kay, J.~Carreira, K.~Simonyan, B.~Zhang, C.~Hillier, S.~Vijayanarasimhan,
  F.~Viola, T.~Green, T.~Back, P.~Natsev \emph{et~al.}, ``The kinetics human
  action video dataset,'' \emph{arXiv preprint arXiv:1705.06950}, 2017.

\bibitem{Shahroudy_2016_NTURGBD}
A.~Shahroudy, J.~Liu, T.-T. Ng, and G.~Wang, ``Ntu rgb+d: A large scale dataset
  for 3d human activity analysis,'' in \emph{IEEE Conference on Computer Vision
  and Pattern Recognition}, 2016.

\bibitem{xiu2018pose}
Y.~Xiu, J.~Li, H.~Wang, Y.~Fang, and C.~Lu, ``Pose flow: Efficient online pose
  tracking,'' \emph{arXiv preprint arXiv:1802.00977}, 2018.

\bibitem{li2017person}
W.~Li, X.~Zhu, and S.~Gong, ``Person re-identification by deep joint learning
  of multi-loss classification,'' \emph{arXiv preprint arXiv:1705.04724}, 2017.

\bibitem{stergiou2020learning}
A.~Stergiou, R.~Poppe, and R.~C. Veltkamp, ``Learning class regularized
  features for action recognition,'' \emph{arXiv preprint arXiv:2002.02651},
  2020.

\bibitem{li2017adaptive}
W.~Li, L.~Wen, M.-C. Chang, S.~Nam~Lim, and S.~Lyu, ``Adaptive rnn tree for
  large-scale human action recognition,'' in \emph{Proceedings of the IEEE
  International Conference on Computer Vision}, 2017, pp. 1444--1452.

\bibitem{weng2017spatio}
J.~Weng, C.~Weng, and J.~Yuan, ``Spatio-temporal naive-bayes nearest-neighbor
  (st-nbnn) for skeleton-based action recognition,'' in \emph{Proceedings of
  the IEEE Conference on Computer Vision and Pattern Recognition}, 2017, pp.
  4171--4180.

\bibitem{duvenaud2015convolutional}
D.~K. Duvenaud, D.~Maclaurin, J.~Iparraguirre, R.~Bombarell, T.~Hirzel,
  A.~Aspuru-Guzik, and R.~P. Adams, ``Convolutional networks on graphs for
  learning molecular fingerprints,'' in \emph{Advances in neural information
  processing systems}, 2015, pp. 2224--2232.

\bibitem{shi2020skeleton}
L.~Shi, Y.~Zhang, J.~Cheng, and H.~Lu, ``Skeleton-based action recognition with
  multi-stream adaptive graph convolutional networks,'' \emph{IEEE Transactions
  on Image Processing}, vol.~29, pp. 9532--9545, 2020.

\bibitem{grover2016node2vec}
A.~Grover and J.~Leskovec, ``node2vec: Scalable feature learning for
  networks,'' in \emph{Proceedings of the 22nd ACM SIGKDD international
  conference on Knowledge discovery and data mining}, 2016, pp. 855--864.

\bibitem{tang2015line}
J.~Tang, M.~Qu, M.~Wang, M.~Zhang, J.~Yan, and Q.~Mei, ``Line: Large-scale
  information network embedding,'' in \emph{Proceedings of the 24th
  international conference on world wide web}, 2015, pp. 1067--1077.

\bibitem{wang2016structural}
D.~Wang, P.~Cui, and W.~Zhu, ``Structural deep network embedding,'' in
  \emph{Proceedings of the 22nd ACM SIGKDD international conference on
  Knowledge discovery and data mining}, 2016, pp. 1225--1234.

\bibitem{cao2015grarep}
S.~Cao, W.~Lu, and Q.~Xu, ``Grarep: Learning graph representations with global
  structural information,'' in \emph{Proceedings of the 24th ACM international
  on conference on information and knowledge management}, 2015, pp. 891--900.

\bibitem{yang2018henneberg}
D.~Yang, M.~Liu, Y.~Zhang, D.~Lin, Z.~Fan, and G.~Chen, ``Henneberg growth of
  social networks: modeling the facebook,'' \emph{IEEE Transactions on Network
  Science and Engineering}, vol.~7, no.~2, pp. 701--712, 2018.

\bibitem{lou2022controllability}
Y.~Lou, D.~Yang, L.~Wang, C.-B. Tang, and G.~Chen, ``Controllability robustness
  of henneberg-growth complex networks,'' \emph{IEEE Access}, 2022.

\bibitem{hu2015jointly}
J.-F. Hu, W.-S. Zheng, J.~Lai, and J.~Zhang, ``Jointly learning heterogeneous
  features for rgb-d activity recognition,'' in \emph{Proceedings of the IEEE
  conference on computer vision and pattern recognition}, 2015, pp. 5344--5352.

\bibitem{liu2017two}
H.~Liu, J.~Tu, and M.~Liu, ``Two-stream 3d convolutional neural network for
  skeleton-based action recognition,'' \emph{arXiv preprint arXiv:1705.08106},
  2017.

\bibitem{liu2016spatio}
J.~Liu, A.~Shahroudy, D.~Xu, and G.~Wang, ``Spatio-temporal lstm with trust
  gates for 3d human action recognition,'' in \emph{European conference on
  computer vision}.\hskip 1em plus 0.5em minus 0.4em\relax Springer, 2016, pp.
  816--833.

\bibitem{zhang2017view}
P.~Zhang, C.~Lan, J.~Xing, W.~Zeng, J.~Xue, and N.~Zheng, ``View adaptive
  recurrent neural networks for high performance human action recognition from
  skeleton data,'' in \emph{Proceedings of the IEEE International Conference on
  Computer Vision}, 2017, pp. 2117--2126.

\bibitem{li2018skeleton}
L.~Li, W.~Zheng, Z.~Zhang, Y.~Huang, and L.~Wang, ``Skeleton-based relational
  modeling for action recognition,'' \emph{arXiv preprint arXiv:1805.02556},
  vol.~1, no.~2, p.~3, 2018.

\bibitem{li2018independently}
S.~Li, W.~Li, C.~Cook, C.~Zhu, and Y.~Gao, ``Independently recurrent neural
  network (indrnn): Building a longer and deeper rnn,'' in \emph{Proceedings of
  the IEEE conference on Computer Vision and Pattern Recognition}, 2018, pp.
  5457--5466.

\bibitem{ke2017new}
Q.~Ke, M.~Bennamoun, S.~An, F.~Sohel, and F.~Boussaid, ``A new representation
  of skeleton sequences for 3d action recognition,'' in \emph{Proceedings of
  the IEEE conference on Computer Vision and Pattern Recognition}, 2017, pp.
  3288--3297.

\bibitem{liu2017enhanced}
M.~Liu, H.~Liu, and C.~Chen, ``Enhanced skeleton visualization for view
  invariant human action recognition,'' \emph{Pattern Recognition}, vol.~68,
  pp. 346--362, 2017.

\bibitem{du2015skeleton}
Y.~Du, Y.~Fu, and L.~Wang, ``Skeleton based action recognition with
  convolutional neural network,'' in \emph{2015 3rd IAPR Asian Conference on
  Pattern Recognition (ACPR)}.\hskip 1em plus 0.5em minus 0.4em\relax IEEE,
  2015, pp. 579--583.

\bibitem{li2017skeleton}
B.~Li, Y.~Dai, X.~Cheng, H.~Chen, Y.~Lin, and M.~He, ``Skeleton based action
  recognition using translation-scale invariant image mapping and multi-scale
  deep cnn,'' in \emph{2017 IEEE International Conference on Multimedia \& Expo
  Workshops (ICMEW)}.\hskip 1em plus 0.5em minus 0.4em\relax IEEE, 2017, pp.
  601--604.

\bibitem{tang2018deep}
Y.~Tang, Y.~Tian, J.~Lu, P.~Li, and J.~Zhou, ``Deep progressive reinforcement
  learning for skeleton-based action recognition,'' in \emph{Proceedings of the
  IEEE Conference on Computer Vision and Pattern Recognition}, 2018, pp.
  5323--5332.

\bibitem{shi2019two}
L.~Shi, Y.~Zhang, J.~Cheng, and H.~Lu, ``Two-stream adaptive graph
  convolutional networks for skeleton-based action recognition,'' in
  \emph{Proceedings of the IEEE Conference on Computer Vision and Pattern
  Recognition}, 2019, pp. 12\,026--12\,035.

\bibitem{fernando2015modeling}
B.~Fernando, E.~Gavves, J.~M. Oramas, A.~Ghodrati, and T.~Tuytelaars,
  ``Modeling video evolution for action recognition,'' in \emph{Proceedings of
  the IEEE Conference on Computer Vision and Pattern Recognition}, 2015, pp.
  5378--5387.

\bibitem{Li_2019_CVPR}
M.~Li, S.~Chen, X.~Chen, Y.~Zhang, Y.~Wang, and Q.~Tian, ``Actional-structural
  graph convolutional networks for skeleton-based action recognition,'' in
  \emph{The IEEE Conference on Computer Vision and Pattern Recognition (CVPR)},
  June 2019.

\bibitem{zhang2019view}
P.~Zhang, C.~Lan, J.~Xing, W.~Zeng, J.~Xue, and N.~Zheng, ``View adaptive
  neural networks for high performance skeleton-based human action
  recognition,'' \emph{IEEE transactions on pattern analysis and machine
  intelligence}, vol.~41, no.~8, pp. 1963--1978, 2019.

\bibitem{zhang2018adding}
P.~Zhang, J.~Xue, C.~Lan, W.~Zeng, Z.~Gao, and N.~Zheng, ``Adding attentiveness
  to the neurons in recurrent neural networks,'' in \emph{proceedings of the
  European conference on computer vision (ECCV)}, 2018, pp. 135--151.

\bibitem{zhang2020semantics}
P.~Zhang, C.~Lan, W.~Zeng, J.~Xing, J.~Xue, and N.~Zheng, ``Semantics-guided
  neural networks for efficient skeleton-based human action recognition,'' in
  \emph{proceedings of the IEEE/CVF conference on computer vision and pattern
  recognition}, 2020, pp. 1112--1121.

\bibitem{hu2019joint}
G.~Hu, B.~Cui, and S.~Yu, ``Joint learning in the spatio-temporal and frequency
  domains for skeleton-based action recognition,'' \emph{IEEE Transactions on
  Multimedia}, vol.~22, no.~9, pp. 2207--2220, 2019.

\end{thebibliography}

%

\begin{IEEEbiography}{Dong Yang}
Biography text here.
\end{IEEEbiography}

\begin{IEEEbiographynophoto}{Mengqi Li}
Biography text here.
\end{IEEEbiographynophoto}


\begin{IEEEbiographynophoto}{Hong Fu}
Biography text here.
\end{IEEEbiographynophoto}

\begin{IEEEbiographynophoto}{Jicong Fan}
Biography text here.
\end{IEEEbiographynophoto}

\begin{IEEEbiographynophoto}{Zhao Zhang}
Biography text here.
\end{IEEEbiographynophoto}




\end{document}